\documentclass[10pt,twocolumn,letterpaper]{article}
\usepackage{cvpr}              %
\usepackage{bm}
\usepackage[accsupp]{axessibility}
\usepackage[dvipsnames]{xcolor}

\usepackage{graphicx}
\usepackage{amsmath}
\usepackage{amssymb}
\usepackage{booktabs}
\usepackage{color}

\renewcommand*{\ie}{i.e.,\@\xspace}
\renewcommand*{\eg}{e.g.,\@\xspace}

\renewcommand{\paragraph}[1]{\vspace{2pt}\noindent\textbf{#1}\hspace{5pt}}

\definecolor{ms_note}{RGB}{0, 181, 190}

\usepackage{pifont}
\definecolor{tickred}{rgb}{0.77, 0.12, 0.23}
\definecolor{tickgreen}{rgb}{0.0, 0.65, 0.31}

\newcommand{\testplanes}{test$^\text{planes}$}
\newcommand{\valplanes}{val$^\text{planes}$}
\DeclareMathOperator{\emb}{\mathbf{e}}
\DeclareMathOperator{\bldp}{\mathbf{p}}
\DeclareMathOperator{\bldx}{\mathbf{x}}
\DeclareMathOperator{\gtpln}{\mathbf{q}}
\DeclareMathOperator{\bldn}{\mathbf{n}}
\DeclareMathOperator{\mlp}{\bm{\phi}}

\graphicspath{{figs/qual_results}{figs/qual_results2}}

\definecolor{cvprblue}{rgb}{0.21,0.49,0.74}
\usepackage[pagebackref,breaklinks,colorlinks,citecolor=cvprblue]{hyperref}

\usepackage{soul}

\def\paperID{12582} %
\def\confName{CVPR}
\def\confYear{2024}

\title{AirPlanes: Accurate Plane Estimation via 3D-Consistent Embeddings}

\newcommand{\gap}{\hspace{20pt}}
\author{Jamie Watson$^{1,3}$ \gap Filippo Aleotti$^{1}$ \gap Mohamed Sayed$^{1}$ \gap Zawar Qureshi$^{1}$ \\ Oisin Mac Aodha$^{2}$ \gap 
 Gabriel Brostow$^{1,3}$ \gap Michael Firman$^{1}$ \gap Sara Vicente$^{1}$ \\ 
 $^{1}$Niantic \hspace{30pt} $^{2}$University of Edinburgh   \hspace{30pt}$^{3}$UCL \\
 \url{https://nianticlabs.github.io/airplanes/}}

\begin{document}
\maketitle

\begin{abstract}
Extracting planes from a 3D scene is useful for downstream tasks in robotics and augmented reality. In this paper we tackle the problem of estimating the planar surfaces in a scene from posed images. Our first finding is that a surprisingly competitive baseline results from combining popular clustering algorithms with recent improvements in 3D geometry estimation. However, such purely geometric methods are understandably oblivious to plane semantics, which are crucial to discerning distinct planes. To overcome this limitation, we propose a method that predicts multi-view consistent plane embeddings that complement geometry when clustering points into planes. We show through extensive evaluation on the ScanNetV2 dataset that our new method outperforms existing approaches and our strong geometric baseline for the task of plane estimation. 
\end{abstract}

\section{Introduction}
\label{sec:intro}

While only parts of the real world are perfectly planar, a 3D reconstruction made out of planes is a useful parameterization for many downstream tasks. 
A planar scene reconstruction is a common representation for applications in robotics~\cite{pathak2010online,geneva2018lips}, path planning~\cite{hummel2006vision}, and augmented reality (AR)~\cite{uematsu2009multiple}. 
For example, both ARKit~\cite{arkit} and ARCore~\cite{arcore}, two of the most used AR platforms, provide 3D plane estimation from scenes as part of their frameworks.

Broadly, there are two families of approaches for 3D plane extraction from images: geometric versus  learning-based methods. 
Geometry-based pipelines assume access to a point cloud or mesh of the scene, \eg as estimated from multi-view stereo or LIDAR. %
This geometry is then partitioned into planes using geometric cues, \eg using RANSAC~\cite{fishler1981ransac}. 
The disadvantages of these approaches are that they can be sensitive to noisy data and they do not typically encode learned priors to facilitate robust plane estimation. 
In contrast, learning-based methods make use of supervised data to develop models that can predict plane parameters from raw images. 
Many prior works have focused on the task of extracting planes from \emph{single} input images~\cite{liu2018planenet,yang2018recovering,liu2019planercnn,yu2019single,tan2021planeTR,shi2023planerectr}. 
In practice though, it is more common to have a \emph{sequence} of input images of the scene of interest, \eg in AR applications where the user is interacting with new parts of the scene in real-time. 
There is however, only limited work that extends these learning-based single image methods to the multi-image setting~\cite{xie2022planarrecon}.  

\begin{figure}[t]
    \centering
    \includegraphics[width=\columnwidth]{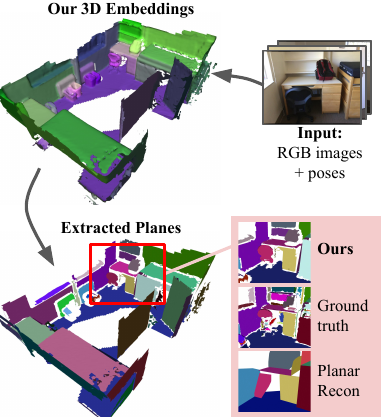}
    \vspace{-15pt}
    \caption{
        \textbf{We create planar scene representations using only posed RGB images as input}.  
        Existing systems can predict \emph{per-pixel} planar embeddings for each image, but these are not 3D consistent. 
        We learn a per-scene function which maps points on the same plane to nearby positions in an embedding space.
        Clustering these embeddings, using strong geometrical priors, gives accurate planar reconstructions.
    }
    \label{fig:intro_fig}
    \vspace{-10pt}
\end{figure}

Inspired by recent work in interactive labeling \cite{zhi2021ilabel}, we propose an alternative approach to discovering planes in 3D.  
We train a small MLP network for each scene, which maps any 3D location in that scene to an embedding vector.
Using various 2D and 3D cues, we train the MLP to produce embeddings which are 3D consistent and can be easily clustered to uncover distinct and accurate planar regions.
By exploiting learned cues when decomposing a scene into planes, our method can adapt to different definitions of \emph{what} constitutes a plane. 
This is important because the concept of what counts as a plane is application dependent. 
For example, a painting on the wall can be considered either a distinct plane or part of the wall plane, depending on the application. 
Unlike purely geometric definitions, our method learns what is considered a plane based on what is ``encoded'' in the training data.

\textbf{Our core contribution} is a new method that estimates 3D-consistent plane embeddings from a sequence of posed RGB images, and then groups them into planar instances. 
We demonstrate via extensive evaluation that our method is more accurate than recent end-to-end learning-based approaches, and can run at interactive speeds.
We also make a surprising observation by proposing an additional strong `geometry plus RANSAC' baseline. 
It can achieve impressive accuracy, outperforming existing baselines, ranking second place behind our proposed method.

\section{Related work}
\label{sec:related_work}

\noindent{\bf Planes from single images.}
Although estimating 3D planes from single images is an ill-posed problem, multiple deep learning solutions have been proposed. 
Top-down approaches~\cite{yang2018recovering,liu2018planenet,liu2019planercnn} directly predict a mask and the parameters of each plane. 
In contrast, bottom-up approaches~\cite{yu2019single} first map pixels into embeddings, which can subsequently be clustered into planes (\eg via clustering methods~\cite{comaniciu2002meanshift}). 
More recent works~\cite{tan2021planeTR,shi2023planerectr} leverage the query learning mechanism of Vision Transformers~\cite{dosovitskiy2020image} to achieve \sota{} single-image results. 
These methods process frames independently, so are unable to produce temporally and 3D-consistent planes. 
As a result, they would require non-trivial plane tracking mechanisms to match the same plane across different frames over time. 
In contrast, we leverage multi-view image sequences, which enables planes to be estimated in 3D rather than just from single images.

We note that some works use planarity assumptions to regularize depth maps~\cite{shao2023nddepth,lee2019big,yu2020p,gallup2010piecewise}
or to improve 3D scenes~\cite{arndt2023planar,ye2023self} and poses~\cite{tan2023nope}.
In contrast, our aim is to find a high quality planar decomposition of the scene, rather than to use planarity for regularization in downstream tasks.

\noindent{\bf Planes from 3D and multi-view images.} 
The extraction of geometric primitives, such as planes, from 3D point clouds is an established problem~\cite{xia2020geometric}.
RANSAC~\cite{fishler1981ransac} and the Hough transform~\cite{hough1962method} are popular strategies to help fit planes, and other 3D shapes~\cite{schnabel2007efficient,borrmann20113d_hough,kluger2021cuboids,ramamonjisoa2022monteboxfinder}, to 3D data.

While a small number of works start from multi-view stereo estimated point clouds~\cite{lafarge2012hybrid,chauve2010robust,bodis2014fast}, the vast majority of plane extraction methods assume access to higher-quality 3D LIDAR  scans~\cite{nan2017polyfit,lafarge2013surface,monszpart2015rapter,chen2008architectural,turner2013watertight}. 
These methods can be slow, \ie not suitable for real-time AR applications, and they cannot easily cope with non-trivial amounts of noise in the input point clouds.  
To address noise, existing methods have attempted to enforce simple to define priors during reconstruction such as a Manhattan-world assumption~\cite{vanegas2012automatic}, object/scene symmetry~\cite{mitra2013symmetry}, or via user interaction~\cite{sinha2008interactive}. 
Methods that only use geometry are fundamentally limited by the quality of the 3D information provided to them. 
In contrast, learning-based methods can learn to compensate for such issues and can also generate planar decompositions that better align with the semantic content of the scene.

Learning-based methods have been proposed for estimating planes from a limited number of input images~\cite{jin2021planar,agarwala2022planeformers,liu2022planemvs}. 
However, extending these methods to entire videos is not trivial. 
Most related to us, PlanarRecon~\cite{xie2022planarrecon} is one of the first learning-based methods to predict a planar representation of entire 3D scenes. 
They incrementally detect and reconstruct 3D planes from posed RGB sequences, where 3D planes are detected in video fragments before the  fragments are fused into a consistent planar reconstruction. 
The pipeline is somewhat complex and contains expensive operations such as 3D convolutions, recurrent units, and differentiable matching. 
In contrast, we trade such complexity for an efficient non-plane-based 3D scene reconstruction method~\cite{sayed2022simplerecon}, which provides reliable scene geometry estimates suitable for input to our plane estimation method. 
Finally, room layout estimation can also leverage multiple images~\cite{hu2022mvlayoutnet,su2023gpr,rozumnyi2023estimating}. 
However, their extreme scene simplification is only suitable for a limited number of applications.

\noindent{\bf 2D and 3D segmentation.}
\label{2d-3d-seg}
Our task of dividing a 3D scene into planes has some similarities with  3D semantic~\cite{rethage2018fully,choy20194d} or panoptic instance~\cite{narita2019panopticfusion,schmid2022panoptic,panopticndt2023iros} segmentation. 
These methods are less applicable to our problem because they aim to segment objects or semantic regions without special regard for geometric properties.
Recent works have leveraged NeRFs to obtain a consistent semantic~\cite{zhi2021inplace} or panoptic~\cite{kunduCVPR2022PNF,siddiqui2023panoptic_lift} scene representation. 
Our method follows this direction by also using test time optimization. 
However, unlike \cite{siddiqui2023panoptic_lift} we use an online rather than offline reconstruction method, and do not need to perform linear assignment for every frame.

\noindent{\bf Scene-level embeddings.}
Our key innovation is to use per-scene 3D embeddings to represent planes.
We are inspired by previous works, \eg iMap~\cite{sucar2021imap} and iLabel~\cite{zhi2021ilabel}, who showed how emergent embeddings can be used for interactive reconstruction and labeling.
We are inspired by these works, but instead of encoding scene geometry or semantic labels, we encode plane embeddings trained to be multi-view consistent. 
Related, there are works that optimize 3D embeddings from 2D supervision, \eg to ground 2D vision-language features~\cite{radford2021learning} in 3D~\cite{kobayashi2022decomposing,kerr2023lerf,tsagkas2023vl,peng2023openscene}. 
However, unlike our reconstruction focus, their aim is to ground open-vocabulary semantic queries in 3D.

\noindent{\bf Representations for 3D reconstruction.}
\label{related-3drec} 
The focus of our work is planar scene representations, but there are many alternatives to planes.
For example, TSDFs encode shape volumetrically. 
They can be generated by estimating depth, \eg from multi-view stereo~\cite{collins1996space,gallup2007real, huang2018deepmvs,duzceker2021deepvideomvs,sayed2022simplerecon}, or directly via more expensive 3D convolutions~\cite{murez2020atlas}.
Subsequent methods~\cite{sun2021neuralrecon,bozic2021transformerfusion} have extended neural TSDF estimation to the online setting. 
Implicit functions are an alternative representation which have been used to map from 3D space to occupancy \cite{liu2019learning,park2019deepsdf,mescheder2019occupancy,runz2020frodo}.
In the context of SLAM, implicit neural strategies have been developed~\cite{sucar2021imap,zhu2022niceslam,zhang2023goslam} that are able to encode scene geometry using a multi-layer perceptron (MLP). 
Finally, further from our task, the recent success of NeRFs~\cite{mildenhall2020nerf} for realistic novel view synthesis has paved the way for methods that apply volume rendering to represent a scene using a neural network~\cite{oechsle2021unisurf,wang2021neus,yu2022monosdf,li2023neuralangelo}.

\section{Method}
\label{sec:method}

We take as input a sequence of color images%
, each associated with a known camera pose.
We aim to predict a representation of the imaged 3D scene, where surfaces are segmented into constituent planes.
We follow the definition of planes from previous work~\cite{liu2019planercnn, liu2022planemvs, xie2022planarrecon} where there can be semantic separation between parts, \eg nearby table-tops should each have a different plane and a closed door should have a different plane to the wall enclosing it.

\begin{figure*}[t]
    \centering
    \includegraphics[width=1\linewidth]{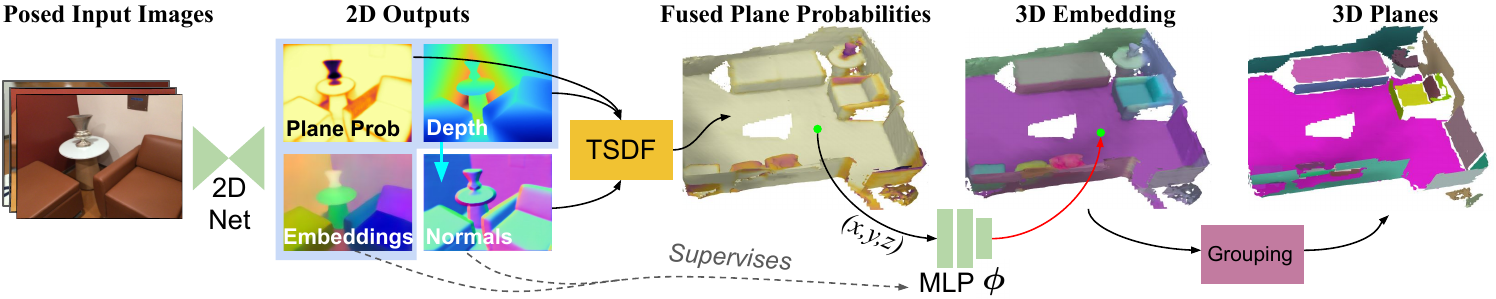}
    \vspace{-15pt}
    \caption{
        \textbf{Our method for 3D plane estimation.} 
        For each RGB keyframe we estimate per-pixel depth, planar probability and planar embedding following \cite{yu2019single}.
        We fuse the depths and planar probabilities into a TSDF and extract a mesh.
        We then train a per-scene MLP to distill the per-pixel embeddings into 3D-consistent embeddings.
        These are finally grouped via clustering into 3D planes. 
    }
    \vspace{-5pt}
    \label{fig:method-overview}
\end{figure*}

Our approach estimates planes by first reconstructing the 3D geometry of the scene using a mesh representation.
We then train a network that maps each point on the mesh to a 3D-consistent embedding space, such that points on the same plane map to nearby places in the space. %
These embeddings \textit{implicitly} encode  semantic instance information and geometric cues.
Semantics complement the 3D geometry information provided by the mesh computed via a lightweight multi-view stereo system. 
We then use a clustering algorithm on the geometry and embeddings to compute accurate plane assignments. %
All the steps in our method support online inference. %
An overview of our approach is shown in Fig.~\ref{fig:method-overview}.

\subsection{Learning 3D planar embeddings}\label{sec:3d_embeddings}
Our key innovation is to learn a mapping from each 3D point $\mathbf{p}$ in a reconstructed scene to an embedding $\emb_{\mathbf{p}}$, such that points on the same plane map to nearly the same place in embedding space, while points on different planes map to different places. 
We denote these as `3D embeddings', where 3D refers to the fact that the embeddings encode \emph{per-scene}, and not \emph{per-image}, planar information. 
We first review how existing \emph{single image} embedding networks are trained, before describing how we distill these pixel-wise embeddings into a 3D-consistent embedding.

\begin{figure}
    \centering
    \includegraphics[width=1.0\columnwidth]{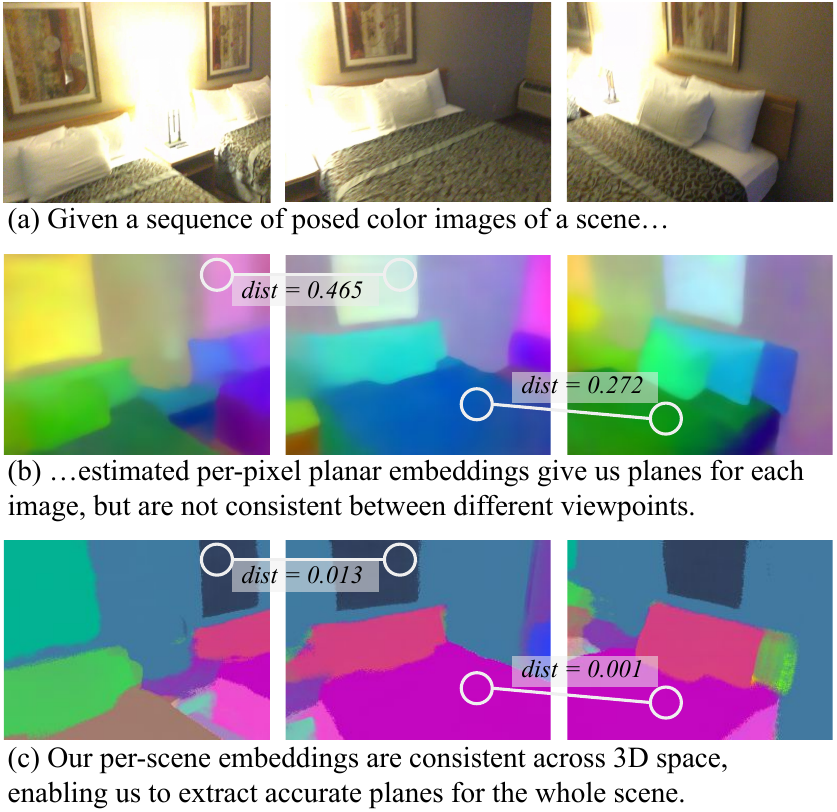}
    \vspace{-19pt}
    \caption{
    \textbf{Per-image planar embeddings are not temporally consistent.} 
    While they can segment planes within a single image, plane embeddings in (b) from~\cite{yu2019single} do not result in 3D consistent embeddings for a full scene. 
    Our method (c) gives a per-scene embedding which is consistent across many views of that scene.}
    \label{fig:embeddings-motivation}
    \vspace{-10pt}
\end{figure}

\paragraph{Single image embeddings.}
In the case of plane estimation from monocular images, \cite{yu2019single} train a feedforward network to map a single color image to per-pixel embeddings.
Pixels $i$ and $j$ in the same image are mapped to embeddings $\bldx_i$ and $\bldx_j$ respectively, where $\bldx_i$ is similar to $\bldx_j$, if and only if $i$ and $j$ are in the same plane. 
This is achieved by training a network which takes as input a \emph{single} image and outputs a per-pixel embedding, using two losses: a \emph{pull} loss penalizing pixel embeddings $\bldx_i$ that are different from the mean embedding of their corresponding plane; and a \emph{push} loss encouraging mean embeddings for each plane to be different from each other. 
One option to obtain 3D embeddings could be to find all pixels that correspond to the reprojection of a 3D point across multiple views and average their per-pixel embeddings. 
The issue with this approach can be seen in Fig.~\ref{fig:embeddings-motivation}. 
Here, the per-pixel embeddings are not consistent across views, despite encoding valuable planar instance information for each individual view.
This is ablated in Sec.~\ref{sec:ablations} as `embeddings w/o test-time optimization'.

\paragraph{Consistent 3D embeddings.}
Our goal is to learn embeddings that preserve the properties of the per-pixel embeddings, while being consistent across views.
We achieve this goal by learning a \emph{per-scene} mapping function $\mlp$, which is parameterized as an MLP and is optimized at test time, following recent work~\cite{zhi2021ilabel,peng2023openscene}.
Our network takes as input a 3D point $\mathbf{p}$ and predicts its `3D' embedding $\emb_{\mathbf{p}} = \mlp(\mathbf{p})$. 

\paragraph{Single image embeddings distillation loss.}
Our network $\mlp$ is trained to distill information contained in the per-pixel embeddings $\bldx$. For a pair of pixels $i$ and $j$ in a single image, we take their embeddings $\bldx_i$ and $\bldx_j$. We also know their corresponding 3D positions $\bldp_i$ and $\bldp_j$ and their image-space normals $\bldn_i$ and $\bldn_j$. 
We can then train the network $\mlp$ such that $\mlp(\bldp_i)$ is similar to $\mlp(\bldp_j)$, if and only if their corresponding embeddings in image space $\bldx_i$ and $\bldx_j$ are similar and their normals ($\bldn_i$ and $\bldn_j$) are also similar.
Inspired by the \emph{push-pull} loss used for the single image embeddings~\cite{yu2019single}, we use the following loss to encourage this:
\begin{equation}
    L_{\mlp} = 
            \begin{cases}
            \lVert \mlp(\bldp_i) - \mlp(\bldp_j) \rVert, \\
            \;\;\;\;\;\;\;\; \text{if} \, \lVert \bldx_i - \bldx_j \rVert  < t_e \, \text{and} \, \bldn_i \cdot \bldn_j > t_n \vspace{10pt}  \\ 
            \max(0, t_p - \lVert \mlp(\bldp_i) - \mlp(\bldp_j)  \rVert), \\
            \;\;\;\;\;\;\;\; \text{otherwise,} \\
            \end{cases}
\label{eqn:mlp_loss}
\end{equation}
where $t_e$ is a \emph{pull} threshold on embeddings, $t_n$ is a threshold on normals, and $t_p$ is a \emph{push} threshold.
This loss is applied to sampled pairs of points on the same image.

\subsection{3D geometry estimation}\label{sec:3d_geometry_estimation}
To estimate planes, we use our 3D embeddings alongside an initial estimate of scene geometry. 
To estimate an accurate 3D mesh we use SimpleRecon~\cite{sayed2022simplerecon}, a state-of-the-art 3D reconstruction system that requires posed images as input. 
In it, depth maps are estimated using a multi-view stereo net, then fused into a 3D mesh via a truncated signed distance function (TSDF)~\cite{curless1996volumetric}. 

We adapt their network to additionally predict a planar/non-planar probability, assigning a per-pixel value indicating if that pixel belongs to a planar or non-planar region, trained equivalently to the single-image plane estimator of~\cite{yu2019single}. 
Our novelty is to then combine these per-pixel predictions into 3D as an additional channel in the TSDF.
When extracting the mesh, we exclude voxels that have an aggregated non-planar value of less than $p = 0.25$, so that non-planar regions are not part of the final mesh.
This extracted mesh is one of the inputs to the next steps.

\subsection{Plane grouping}\label{sec:clustering}

Given an embedding for each vertex in our 3D mesh,  our next step is to cluster vertices into plane instances based on those embeddings and on geometry information defined by the mesh.
For this clustering step we rely on sequential RANSAC \cite{fishler1981ransac}.
RANSAC works by randomly sampling plane instance proposals, checking the inlier count for each proposal, and selecting the plane instance with the most inliers. This process is done sequentially, where at each iteration the points associated with the last predicted plane are removed from the pool.
Each plane instance proposal is created by sampling a single mesh vertex, which together with its associated normal, defines a plane.
A different mesh vertex is considered an inlier to this plane proposal if: (i) the distance to the plane is smaller than a threshold $r_d$ and (ii) the euclidean difference between embeddings is smaller than a threshold $r_e$.
After convergence, we merge planes with highly similar embeddings and normals, \ie where the distance between average embeddings is $< 0.2$ and the dot product between average normals is $> 0.6$.
Next, we run a connected components algorithm on the mesh representation of each discovered plane in turn, to separate out non-contiguous planes. 

Since the non-planar vertices have already been removed as explained in Sec.~\ref{sec:3d_geometry_estimation}, we expect all remaining vertices to be assigned a plane instance label. 
RANSAC, however does not guarantee this. 
For this reason, we run a post-processing step that iteratively propagates labels to connected unlabeled points from the RANSAC step.
Finally, we remove planes with fewer than $100$ vertices.

\subsection{Online inference}\label{sec:online}
\begin{figure}[t]
    \centering
    \includegraphics[width=\columnwidth]{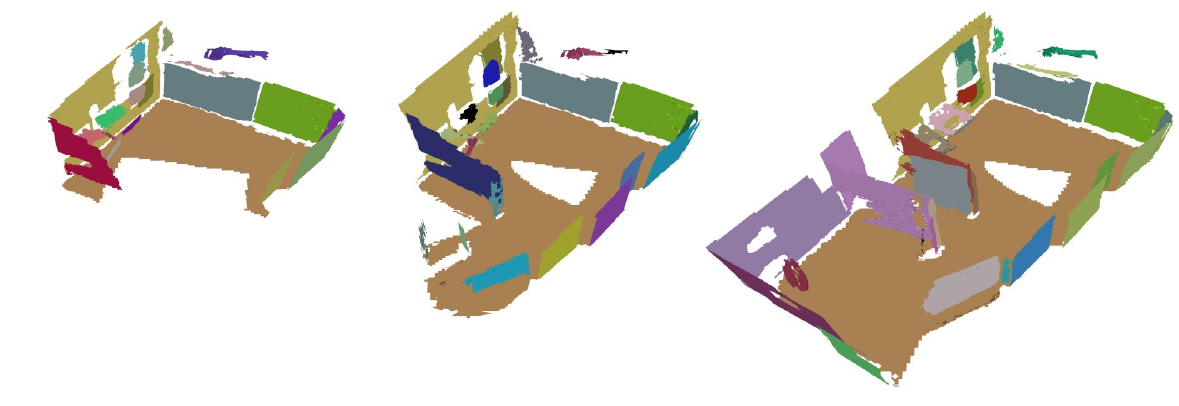} \\
    \vspace{-10pt}
    \caption{
        \textbf{Planes can be estimated online at interactive rates.}
        As new RGB frames are acquired, we can update the weights of our MLP and recompute plane assignments.
        See Sec.~\ref{sec:timings} for timings.
    }
    \label{fig:online}
    \vspace{-10pt}
\end{figure}

All components of our method are designed so that they can run online with little adaptation.
The 3D geometry estimation steps, \ie depth estimation, fusion into TSDF, and mesh extraction, are commonly used in online systems~\cite{newcombe2011kinectfusion, sayed2022simplerecon}. 
Our per-scene embedding network is always updated in an online fashion, similar to~\cite{sucar2021imap, zhi2021ilabel}. 
Given the current 3D mesh and the current embedding network, embeddings can be predicted for all mesh vertices. 
We then perform clustering to extract plane instances.
To achieve interactive speeds for our online method, we replace our RANSAC clustering method, which takes 131ms on average per scene, with the mean-shift algorithm~\cite{comaniciu2002meanshift} using the efficient implementation from~\cite{yu2019single}, which takes 25ms.
We evaluate this alternative clustering algorithm in the experimental section.
Finally, each time we recompute planes, we use Hungarian matching~\cite{kuhn1955hungarian} between the previous and current plane assignments to encourage consistency of planes across time (visible in the figure as stability of colors over time, while new planes are computed).
Fig.~\ref{fig:online} shows an online reconstruction obtained with our method for a ScanNetV2 scene.

\subsection{Sequential RANSAC: A strong baseline} 
\label{ransac}
\begin{figure}[!t]
    \centering
    \includegraphics[width=1.0\columnwidth]{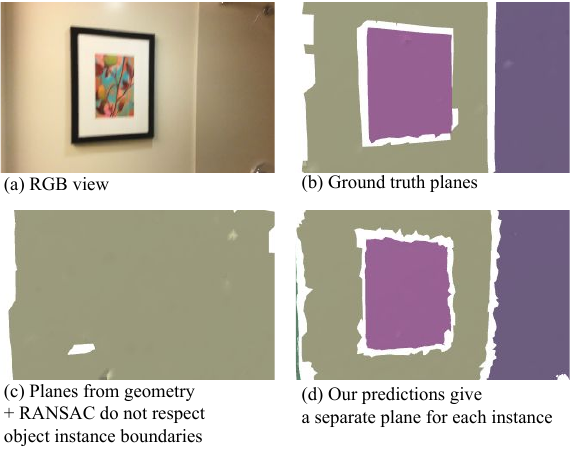}
    \vspace{-19pt}
    \caption{
        \textbf{Sequential RANSAC alone is not enough to segment planar instances.} 
        Sequential RANSAC (with geometry from \cite{sayed2022simplerecon}) does well, but fails to segment adjacent co-planar instances.
        Our method can segment these, \eg this picture frame.
    }
    \label{fig:ransac-problems}
    \vspace{-8pt}
\end{figure}

Given recent advances in 3D scene reconstruction from image inputs, \eg~\cite{sayed2022simplerecon}, the question arises: 
How good a planar decomposition can we achieve if we run RANSAC on the mesh only, without the contribution of our 3D embeddings? 
Surprisingly, later results show this simple baseline performs very well. %
However, while this naive approach takes geometry into account, it does not leverage semantic or appearance-based cues, leading to plane over- and under-segmentation issues (see Fig.~\ref{fig:ransac-problems}). 
Our method, using 3D plane embeddings, addresses these problems.

\begin{table*}
    \renewcommand{\tabcolsep}{10pt}
    \centering
    \small
    \resizebox{1.0\textwidth}{!} 
    {
    \begin{tabular}{|l|cc|ccc|ccc|}
        \hline
            & \multicolumn{2}{c}{Geometry} & \multicolumn{3}{|c|}{Segmentation} & \multicolumn{3}{c|}{Planar} \\
        \hline
            & chamfer ↓  & f1 ↑  & voi↓	& ri↑	& sc↑ & fidelity↓ & accuracy↓ & chamfer↓\\
        \hline
    PlaneRecTR~\cite{shi2023planerectr} + aggregation &8.82 & 42.53 & 4.028 & 0.924 & 0.268 & 22.58 & 15.71 & 19.14 \\
    
    Atlas~\cite{murez2020atlas}~$\dagger$ + RANSAC &12.65 & 53.71 & 2.868 & 0.932 & 0.465 & 22.60 & 17.71 & 20.16 \\
    
    NeuralRecon~\cite{sun2021neuralrecon} + RANSAC &9.00 & 46.91 & 3.176 & 0.929 & 0.391 & 16.76 & 13.08 & 14.92 \\
    
    FineRecon~\cite{stier2023finerecon} + RANSAC &5.56 & 64.10 &  \underline{2.377} &  \underline{0.950} &  \underline{0.531} &  \textbf{7.74} & 11.71 &  \underline{9.72} \\
    
    SR~\cite{sayed2022simplerecon} + RANSAC & \underline{5.40} &  \textbf{65.45} & 2.507 & 0.946 & 0.515 & 9.42 &  \underline{10.13} & 9.78 \\
    
    PlanarRecon~\cite{xie2022planarrecon} &9.89 & 43.47 & 3.201 & 0.919 & 0.405 & 18.86 & 16.21 & 17.53 \\
    
    \textbf{Ours} & \textbf{5.30} &  \underline{64.92} &  \textbf{2.268} &  \textbf{0.957} &  \textbf{0.568} &  \underline{8.76} &  \textbf{7.98} &  \textbf{8.37} \\
        \hline
    \end{tabular}
    }
    \vspace{-5pt}
    \caption{\textbf{Quantitative evaluation.} We report geometry scores using the \testplanes{} split of ScanNetV2~\cite{dai2017scannet}. Here, \textbf{bold} indicates best, \underline{underline} second best.
    We use publicly available checkpoints unless indicated. $\dagger$ indicates that we use fewer voxels to fit in memory.}
    \vspace{-10pt}
    \label{tab:baselines}
\end{table*}

\begin{table}
    \renewcommand{\tabcolsep}{3pt}
    \centering
    \small
     \resizebox{0.48\textwidth}{!} 
    {
    \begin{tabular}{|l|c|c|c|}
        \hline
            & Geometry & Segmentation & Planar \\
            & chamfer↓ & voi↓ & chamfer↓\\
        \hline
        Atlas~\cite{murez2020atlas}~$\dagger$ + RANSAC & 12.65 & 2.868 & 20.16 \\
        \hspace{10 pt} + our 3D embeddings &  \textbf{12.30} &  \textbf{2.673} &  \textbf{18.92} \\
        \hline
        NeuralRecon~\cite{sun2021neuralrecon} + RANSAC & 9.00 & 3.176 & 14.92 \\
        \hspace{10 pt} + our 3D embeddings &  \textbf{8.54} &  \textbf{2.713} &  \textbf{13.42} \\
        \hline
        FineRecon~\cite{stier2023finerecon} + RANSAC & 5.56 & 2.377 & 9.72 \\
        \hspace{10 pt} + our 3D embeddings &  \textbf{4.80} &  \textbf{2.159} &  \textbf{7.76} \\
        \hline
        SR~\cite{sayed2022simplerecon} + RANSAC & 5.40 & 2.507 & 9.78 \\
        \hspace{10 pt} + our 3D embeddings \textbf{(Ours)} &  \textbf{5.30} &  \textbf{2.268} &  \textbf{8.37} \\
        \hline
    \end{tabular}
    }
    \vspace{-5pt}
    \caption{\textbf{Our 3D embeddings} can be used in combination with a variety of different 3D geometry estimators, leading to improved results for all methods.}
    \vspace{-10pt}
    \label{tab:ours_with_different_geometry}
\end{table}

\section{Implementation details}
\label{subsec:implementation_details}

\noindent{\bf Depth, plane probabilities, and per-pixel embedding network architecture.}
We use the SimpleRecon \cite{sayed2022simplerecon} architecture for depth estimation. 
Encoder features are shared between the depth estimation, plane probabilities, and per-pixel embedding tasks, though they have separate decoders. 
Full architecture details are in the supplementary material.

\noindent{\bf Embedding MLP network.} We use a three-layer MLP with 128 dimensions for each hidden layer.
Following \cite{sitzmann2020implicit}, we lift the input to the MLP to 48 periodic activation functions before it is input to the first linear layer.
Our final embedding has three dimensions.
We use $t_e = 0.9$, $t_n = 0.8$ and $t_p = 1.0$, tuned on the validation set.
Similarly to \cite{zhi2021ilabel}, the MLP is always trained in an online fashion.
For each new keyframe we sample 400 pixels from it and apply Eqn.~(\ref{eqn:mlp_loss}) to each pair of points,  together with the pairs from the $10$ most recent keyframes. 
We then run backpropagation ten times to optimize the MLP.

\noindent{\bf Grouping thresholds.} For RANSAC we set $r_e = 0.5$ and $r_d = 0.1$. We set the mean-shift bandwidth to $0.25$.

\noindent{\bf Mesh planarization.}
Given our final assignment of points to planes, we perform \emph{mesh planarization} to convert our 3D mesh into a planarized mesh.
First, we estimate the plane equation for each plane.
Next, each point is moved along the normal of its assigned plane such that it lies on the plane it is assigned to. 
This is the mesh which is geometrically evaluated against the ground truth planarized mesh.

\section{Experiments}
\label{sec:experiments}

We train and evaluate on ScanNetV2~\cite{dai2017scannet}, because~\cite{liu2019planercnn} provided ground truth plane annotations for most of it.
Plane annotations are unavailable for the ScanNetV2 test set.
We therefore split the official  ScanNetV2 validation set into new \emph{plane evaluation} validation and test splits, dubbed \valplanes{} and \testplanes{}, with $80$ and $100$ scenes respectively.
For a fair comparison with prior work, we re-evaluate baselines on our new test split.
The new splits and our evaluation code are available at \href{https://nianticlabs.github.io/airplanes/}{{https://nianticlabs.github.io/airplanes/}}.

\subsection{Evaluation metrics} 

\paragraph{Geometric evaluation.}
Here, we evaluate how well the predicted planar mesh approximates the geometry of the ground truth planar mesh.
Following~\cite{xie2022planarrecon} we adopt conventional 3D metrics \cite{murez2020atlas,bozic2021transformerfusion}.
To compare a predicted mesh with the ground truth mesh, we first sample $N=200,000$ points from each mesh. 
We then compare the two sampled point clouds to each other using chamfer distance and f1 score. See~\cite{bozic2021transformerfusion} for details.

Fully volumetric methods such as \cite{murez2020atlas} predict geometry for the whole scene, including unobserved regions. 
To prevent such methods from being penalized unfairly, we enforce a visibility mask to handle unseen points differently when computing metrics, following~\cite{bozic2021transformerfusion, sayed2022simplerecon}.
This visibility mask is applied to all methods for fair comparison.
We also mask out 3D points sampled on faces that connect two or more planes, as these points have ambiguous labeling.
For full transparency, we report numbers in the supplementary material using the evaluation method  from~\cite{xie2022planarrecon} without our additions.

\paragraph{Plane segmentation evaluation.}
Following previous work on plane estimation \cite{yang2018recovering, yu2019single, shi2023planerectr, xie2022planarrecon}, we also report the following clustering metrics: Variation of Information (VOI), Rand Index (RI), and Segmentation Covering (SC).
Given a predicted mesh, we use the protocol proposed in~\cite{xie2022planarrecon} to map the plane ID of each vertex to the closest vertex in the ground truth mesh. See~\cite{xie2022planarrecon} for full details.

\begin{table*}
    \renewcommand{\tabcolsep}{8pt}
    \centering
    \resizebox{1.0\textwidth}{!}{
    \begin{tabular}{|l|cc|ccc|ccc|}
        \hline
            & \multicolumn{2}{c}{Geometry} & \multicolumn{3}{|c|}{Segmentation} & \multicolumn{3}{c|}{Planar} \\
        \hline
            & chamfer ↓  & f1 ↑  & voi↓	& ri↑	& sc↑ &  fidelity↓ & accuracy↓ & chamfer↓\\
        \hline
        Fused per-pixel embeddings w/o test time optimization & 5.76 & 61.83 & 2.670 & 0.949 & 0.485 & 11.63 & 8.30 & 9.97\\
        Fused per-pixel embeddings w.~train.~time m-v consist. & 5.72 & 61.91 & 2.672 &  0.950 & 0.485 & 11.90 &  8.01 & 9.96\\
Ours without planar probability &5.39 & 64.04 & 2.257 & 0.958 & 0.570 & 8.67 & 8.12 & 8.39 \\
    \hline
        \textbf{Ours} (RANSAC)   & 5.30 & 64.92 &  2.268 &  0.957 &  0.568 &  8.76 & 7.98  & 8.37 \\
        \textbf{Ours} (Mean-shift)  & 5.70 & 62.73 & 2.344 & 0.954 & 0.556 & 9.56 & 8.19 & 8.88 \\
    \hline
    SR~\cite{sayed2022simplerecon} +  RANSAC + predicted semantic labels &5.56 & 64.79 & 2.483 & 0.948 & 0.525 & 10.20 & 7.81 & 9.01 \\
    
    SR~\cite{sayed2022simplerecon} +  RANSAC + g.t.~semantic labels &5.41 & 65.71 & 2.262 & 0.956 & 0.568 & 9.96 & 6.13 & 8.04 \\
    
    SR~\cite{sayed2022simplerecon} + RANSAC + g.t.~instance labels &5.68 & 65.57 & 2.257 & 0.954 & 0.584 & 9.94 & 4.90 & 7.42 \\
    \hline 
    \end{tabular}
    }
    \vspace{-5pt}
    \caption{\textbf{Our contributions result in the best performing model.} In these ablations, we turn parts of our method off in turn, or replace them with alternatives, to show the benefit our contributions bring.
    }
    \vspace{-7pt}
    \label{tab:ablations}
\end{table*}

\paragraph{Planar metrics.} 
To better evaluate how well the main, \ie large planes in the ground truth scene, are reconstructed we additionally propose the following protocol.
We select the $k=20$ largest planes from each ground truth mesh. 
For each such plane $\gtpln_j$, we find the predicted plane $\bldp_i$ that most closely matches according to the completion metric. 
We report the fidelity between $\gtpln_j$ and $\bldp_i$ as $\textit{completion}(\gtpln_j, \bldp_i)$, where $\textit{completion}$ is the completion metric from \cite{murez2020atlas}. 
The average of this score over all $k$ ground truth planes over all scenes is our \textit{planar fidelity} score.
We also report the geometric accuracy between $\gtpln_j$ and $\bldp_i$ as \textit{planar accuracy}, and the average of the two as \textit{planar chamfer}.

\subsection{Comparisons with baselines} 

We evaluate our 3D plane estimation method against various baselines (Table~\ref{tab:baselines}).
\textbf{PlanarRecon}~\cite{xie2022planarrecon} is the existing state-of-the-art method for 3D plane estimation from posed RGB images.
We outperform their approach in geometry, segmentation, and planar metrics.
We also compare with the leading baseline for 3D plane estimation from a single image, \textbf{PlaneRecTR} \cite{shi2023planerectr}. For each scene, we run this single image predictor for selected keyframes. 
Planes from each incoming image are matched to the closest world planes by comparing planar normals, offsets, and plane positions.

We compare with our own implementation of sequential \textbf{RANSAC}, applied to meshes from SimpleRecon~\cite{sayed2022simplerecon}. SimpleRecon (SR in tables) is the same method we use for geometry estimation, as detailed in Sec.~\ref{sec:3d_geometry_estimation}, making this the closest baseline to our method, but without using the benefits of our 3D consistent embeddings.
In addition, we also apply the sequential RANSAC method to geometry from~\cite{murez2020atlas, sun2021neuralrecon, stier2023finerecon}. 
See supplementary material for implementation details of the baselines.

Our method outperforms all other methods on the segmentation metrics. While the results for the geometric metrics are comparable with the \emph{SR \cite{sayed2022simplerecon} + RANSAC} baseline, we significantly outperform this baseline on the segmentation and planar metrics, clearly demonstrating the benefit of using our 3D consistent embeddings. %
Surprisingly, PlanarRecon~\cite{xie2022planarrecon} is outperformed by several of our sequential RANSAC baselines. This is in contrast with the results presented in~\cite{xie2022planarrecon}, and we discuss this difference in more detail in the supplementary material.

\paragraph{Our embeddings benefit other geometry methods.} 
To validate the usefulness of our 3D embeddings, we use them in combination with different geometry estimation methods \cite{murez2020atlas, sun2021neuralrecon, stier2023finerecon, sayed2022simplerecon}. We compare using only 3D geometry versus using 3D geometry plus the embeddings derived from our test-time optimized MLPs without retraining. 
We show the results for this experiment in Table~\ref{tab:ours_with_different_geometry}. For all methods, we observe that the additional information encoded in the embeddings improves over the baseline of using geometry + RANSAC only.

\begin{figure}
    \centering
    \newcommand{\qualimwidth}{0.32\columnwidth}
    \renewcommand{\tabcolsep}{1.5pt}
    \footnotesize
    \begin{tabular}{ccc}
        \centering
        \textbf{RGB Mesh} & \textbf{SR~\cite{sayed2022simplerecon} + RANSAC}  & \textbf{Ours} \\

        \includegraphics[width=\qualimwidth,clip]{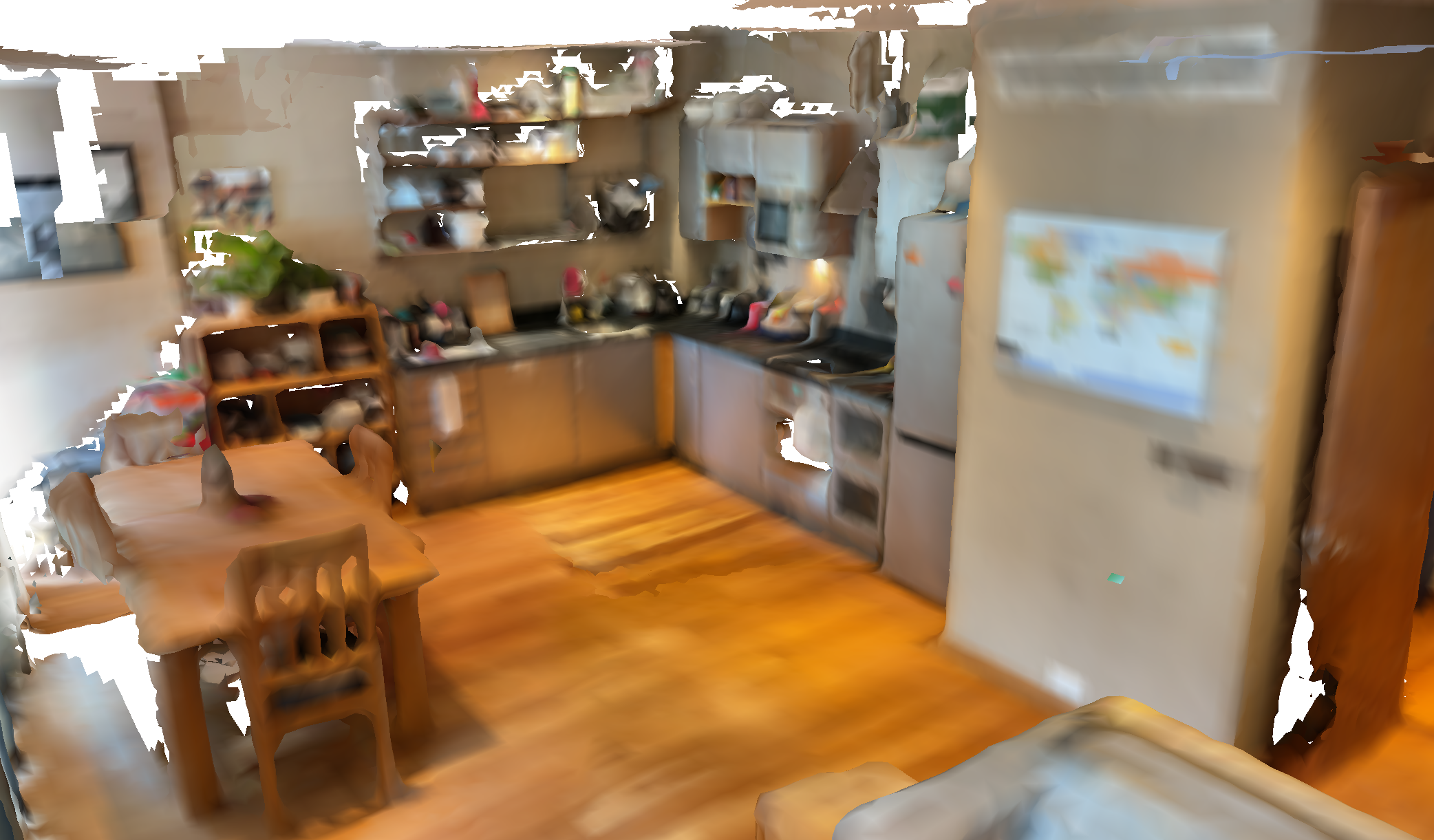} & 
        \includegraphics[width=\qualimwidth,clip]{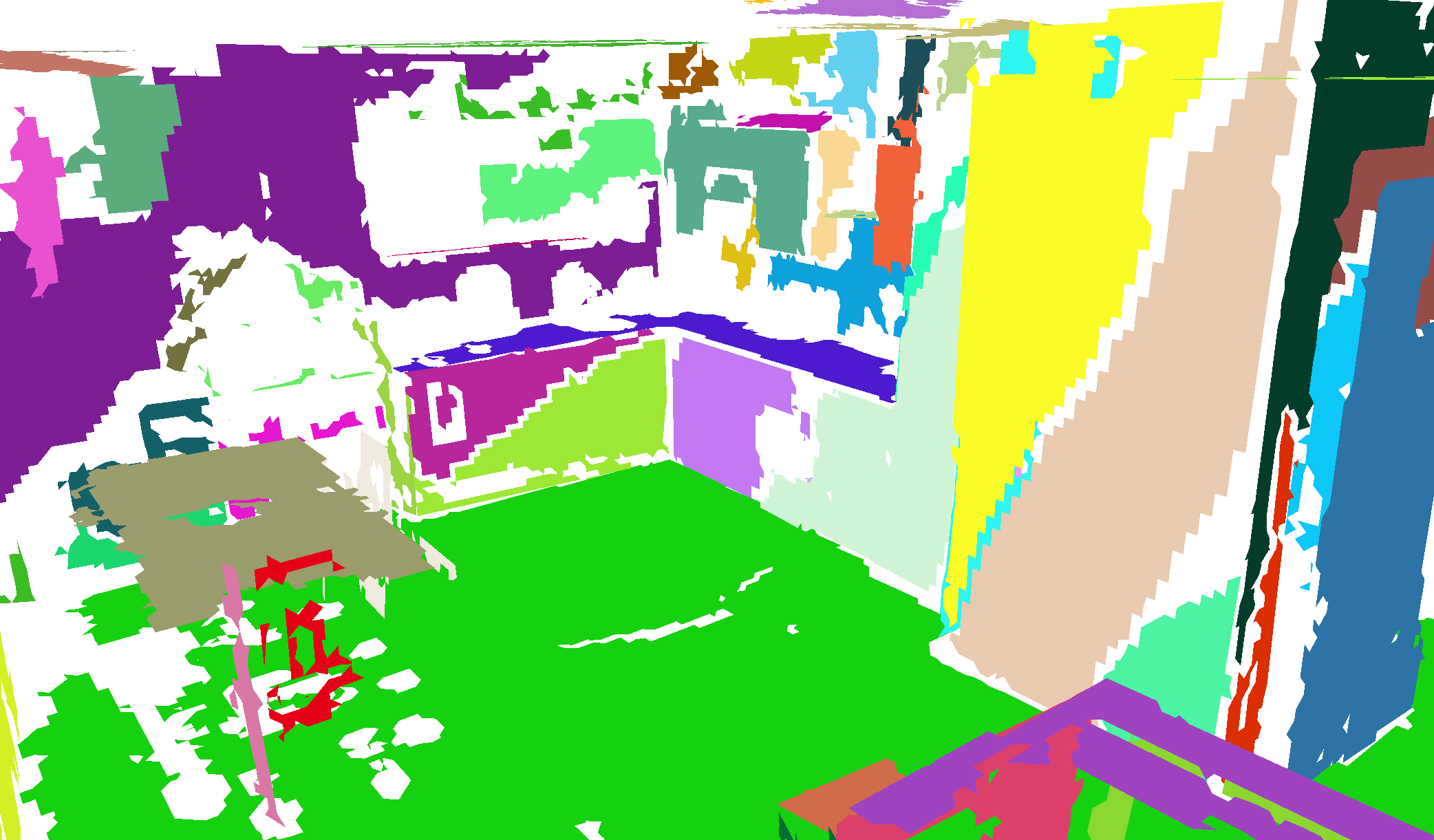} & 
        \includegraphics[width=\qualimwidth,clip]{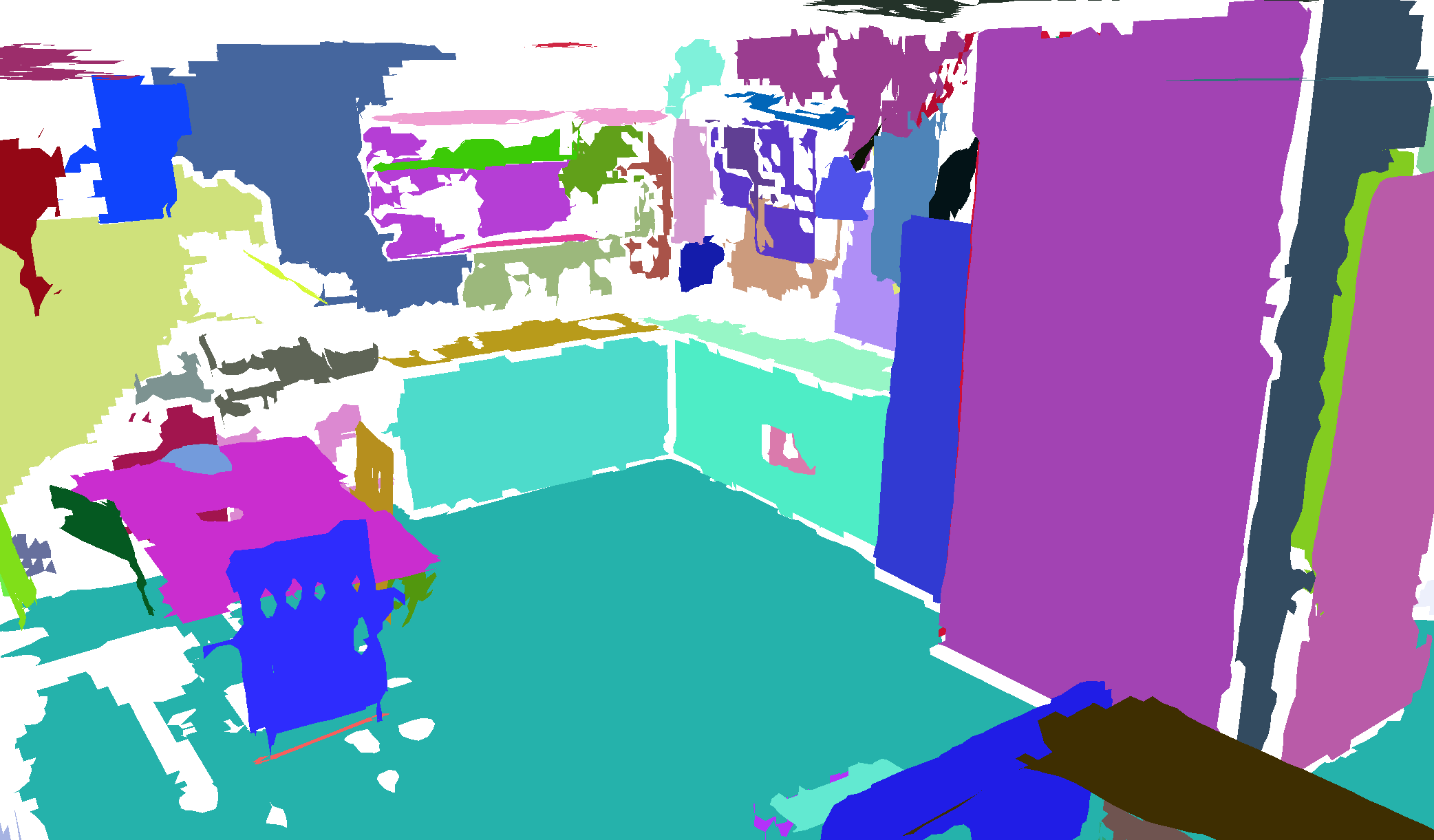} \\
    \end{tabular}
    \vspace{-8pt}
    \caption{\textbf{Our method generalizes well to other scenes.} Here we show results on some non-ScanNetV2 casually-captured footage. 
    }
    \vspace{-12pt}
    \label{fig:custom-scenes}
\end{figure}

\subsection{Ablations}
\label{sec:ablations}

We ablate our method to validate that our contributions lead to higher scores. These results are in Table~\ref{tab:ablations}.
\textbf{Fused per-pixel embeddings w/o test time optimization} is our method but using the embeddings directly from \cite{yu2019single}, without our 3D distillation. These embeddings are fused as additional channels into the TSDF.
\textbf{Fused per-pixel embeddings with training time multi-view consistency} is a variant of our method, where we attempt to train a single feed-forward embedding network which predicts multi-view consistent 3D embeddings directly, without performing test-time optimization.
\textbf{Ours without planar probability} is our method but all points are assigned a planar probability of $1$, meaning non-planar points are still part of the mesh. For this reason, we do not run the post-processing step that assigns unlabeled points after RANSAC.

The first two ablations show that it is not trivial to predict 3D consistent embeddings using a feed-forward network applied to each frame independently. This motivates our use of a per-scene MLP optimized at test time to achieve consistent embeddings. 
The last of these ablations shows that our fusion of planar probabilities into the TSDF improves geometry metrics. 
We note that some computational savings could be made, at the price of  $\sim$1\% drop in geometry scores, without this.

We also compare the two different {plane grouping} algorithms. 
Using the {Mean-shift} variant of our method leads to only a small degradation of results versus RANSAC, while achieving interactive speeds (see Sec.~\ref{sec:timings} for timings).

\textbf{RANSAC oracle methods} are variants of RANSAC which have access to ground truth semantic and instance information.
\textbf{SR + RANSAC + ground truth semantic labels} uses ground truth semantic labels (transferred to the closest vertex in the predicted mesh) in the sequential RANSAC loop to separate planes. Specifically, points can be associated with a plane candidate only if they are geometrically consistent and have the same label.
We additionally compare with a RANSAC variant with \textbf{\emph{predicted} semantic labels}, where we predict $N=20$ semantic classes and we fuse their probabilities in the TSDF. It is worth noting that explicitly predicting semantics is beneficial and leads to better planar scores compared to its geometry-only counterpart in Table~\ref{tab:baselines}. However, our method provides better results across all metrics, and requires fusion of only planar probabilities instead of $N$ semantic classes, which might be challenging as $N$ increases.
Finally, we also show an oracle with \textbf{ground truth instance labels}, which presents an upper bound for plane estimation. 

\begin{figure*}
    \centering
    \vspace{-5pt}
    \newcommand{\qualimwidth}{0.23\textwidth}
    \renewcommand{\tabcolsep}{2pt}
    \small
    \begin{tabular}{cccc}
        \centering
        \textbf{Ground truth} & \textbf{Ours} &  \textbf{SR~\cite{sayed2022simplerecon} + RANSAC} & \textbf{PlanarRecon~\cite{xie2022planarrecon}} \\
        \input{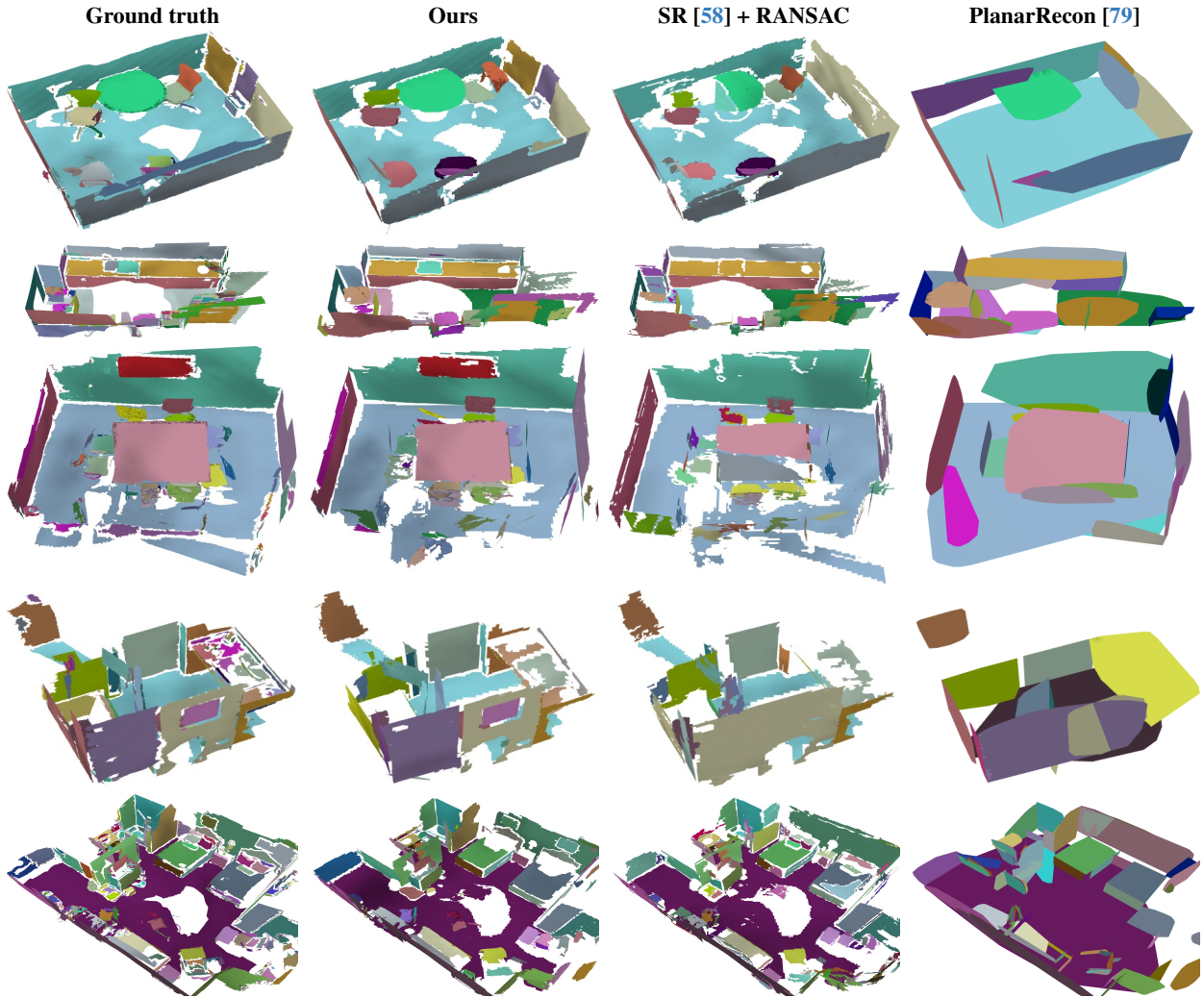}
    \end{tabular}
    \vspace{-10pt}
    \caption{\textbf{Qualitative results on ScanNetV2.} Our predictions are more faithful to the ground truth both in terms of geometry and segmentation. Here we have removed ceiling planes (\ie those with normals facing downwards) for visualization.
    In the first row, we recover the size and shape of the table well when compared to the baselines.
    In the second row, the sink is well separated from the countertop in our results.
    The bottom row shows a failure case where we do not recover small pillows on the beds.
    }
    \vspace{-10pt}
    \label{fig:qualitative-comparison}    
\end{figure*}

\subsection{Qualitative results}

Fig.~\ref{fig:qualitative-comparison} shows results of our method compared to the closest published competitor \emph{PlanarRecon~\cite{xie2022planarrecon}} and our \emph{SR \cite{sayed2022simplerecon} + RANSAC} baseline.
We can see that our method has closer fidelity to the ground truth versus~\cite{xie2022planarrecon}, and avoids oversimplification of geometry.
By more closely adhering to the geometry of the real scene, our planes can appear to have `jagged' edges when compared to the more simplified outputs from~\cite{xie2022planarrecon}. 
Our planar meshes have gaps where planes intersect because we remove triangles that connect vertices from different planes.
If needed for a specific application, our outputs could be further post-processed, \eg using~\cite{nan2017polyfit}. 
A qualitative comparison of our method with the \emph{SR \cite{sayed2022simplerecon} + RANSAC} baseline shows that we are able to recover separate semantic planes that have a common planar geometry.
Finally, Fig.~\ref{fig:custom-scenes} shows more results of our method, with images and camera poses from an iPhone running ARKit~\cite{arkit}.

\subsection{Planes at interactive speeds}
\label{sec:timings}
The online variant of our method, which uses mean-shift clustering, takes a total of 152ms per keyframe on average, on an RTX A6000 GPU. 
This comprises 65ms to obtain the per-pixel depth, planar probability, and 2D planar embedding and separately 1ms for TSDF fusion, 61ms to update the MLP, and 25ms to run the clustering. 
As the average interval between keyframes in ScanNetV2 is 272ms, our method runs at interactive speeds. 
Alternatively, for the RANSAC variant, the clustering step takes 131ms for an entire scene.

\subsection{Limitations}
Our method shows notable improvements compared to other 3D plane estimation methods, but limitations remain. 
Errors in the geometry from our MVS system might have severe consequences when extracting 3D planes.
We also fit planes in a greedy manner. 
Instead, global optimization \eg~\cite{isack2012energy} may further improve results.
Unlike \cite{xie2022planarrecon}, we only estimate planes for visible geometry. Completing unobserved regions, like \cite{murez2020atlas, stier2023finerecon,dai2018scancomplete}, could be a useful extension for some applications.

\section{Conclusion}
\label{sec:conclusion}
We propose a new approach which takes a sequence of posed color images as input, and outputs a planar representation of the 3D scene. 
Surprisingly,  we demonstrate that a strong baseline for this task is to simply run sequential RANSAC on a lightweight 3D reconstruction. 
However, this baseline is likely too limited for AR and robotics use-cases. Our approach addresses this, by training a 3D embedding network to map 3D points to 3D-consistent and meaningful plane embeddings, which can then be clustered into 3D planes.
Our approach gives state-of-the-art plane estimation performance on the ScanNetV2 dataset.

\paragraph{Acknowledgements.} We are extremely grateful to Saki Shinoda, Jakub Powierza, and Stanimir Vichev for their invaluable infrastructure support.

\clearpage
{
    \small
    \bibliographystyle{ieeenat_fullname}
    \bibliography{main}

\begin{thebibliography}{88}
\providecommand{\natexlab}[1]{#1}
\providecommand{\url}[1]{\texttt{#1}}
\expandafter\ifx\csname urlstyle\endcsname\relax
  \providecommand{\doi}[1]{doi: #1}\else
  \providecommand{\doi}{doi: \begingroup \urlstyle{rm}\Url}\fi

\bibitem[Agarwala et~al.(2022)Agarwala, Jin, Rockwell, and
  Fouhey]{agarwala2022planeformers}
Samir Agarwala, Linyi Jin, Chris Rockwell, and David~F Fouhey.
\newblock Planeformers: From sparse view planes to {3D} reconstruction.
\newblock In \emph{ECCV}, 2022.

\bibitem[Apple(2023)]{arkit}
Apple.
\newblock {ARKit}, 2023.
\newblock {Accessed: 5 October 2023}.

\bibitem[Arndt et~al.(2023)Arndt, Sabzevari, and Civera]{arndt2023planar}
Charlotte Arndt, Reza Sabzevari, and Javier Civera.
\newblock Do planar constraints improve camera pose estimation in monocular
  {SLAM}?
\newblock In \emph{ICCV}, 2023.

\bibitem[B{\'o}dis-Szomor{\'u} et~al.(2014)B{\'o}dis-Szomor{\'u},
  Riemenschneider, and Van~Gool]{bodis2014fast}
Andr{\'a}s B{\'o}dis-Szomor{\'u}, Hayko Riemenschneider, and Luc Van~Gool.
\newblock Fast, approximate piecewise-planar modeling based on sparse
  structure-from-motion and superpixels.
\newblock In \emph{CVPR}, 2014.

\bibitem[Borrmann et~al.(2011)Borrmann, Elseberg, Lingemann, and
  N{\"u}chter]{borrmann20113d_hough}
Dorit Borrmann, Jan Elseberg, Kai Lingemann, and Andreas N{\"u}chter.
\newblock The {3D} hough transform for plane detection in point clouds: A
  review and a new accumulator design.
\newblock \emph{3D Research}, 2011.

\bibitem[Bozic et~al.(2021)Bozic, Palafox, Thies, Dai, and
  Nie{\ss}ner]{bozic2021transformerfusion}
Aljaz Bozic, Pablo Palafox, Justus Thies, Angela Dai, and Matthias Nie{\ss}ner.
\newblock {TransformerFusion}: Monocular {RGB} scene reconstruction using
  transformers.
\newblock \emph{NeurIPS}, 2021.

\bibitem[Chauve et~al.(2010)Chauve, Labatut, and Pons]{chauve2010robust}
Anne-Laure Chauve, Patrick Labatut, and Jean-Philippe Pons.
\newblock Robust piecewise-planar {3D} reconstruction and completion from
  large-scale unstructured point data.
\newblock In \emph{CVPR}, 2010.

\bibitem[Chen and Chen(2008)]{chen2008architectural}
Jie Chen and Baoquan Chen.
\newblock Architectural modeling from sparsely scanned range data.
\newblock \emph{IJCV}, 2008.

\bibitem[Choy et~al.(2019)Choy, Gwak, and Savarese]{choy20194d}
Christopher Choy, JunYoung Gwak, and Silvio Savarese.
\newblock {4D} spatio-temporal convnets: Minkowski convolutional neural
  networks.
\newblock In \emph{CVPR}, 2019.

\bibitem[Collins(1996)]{collins1996space}
Robert~T Collins.
\newblock A space-sweep approach to true multi-image matching.
\newblock In \emph{CVPR}, 1996.

\bibitem[Comaniciu and Meer(2002)]{comaniciu2002meanshift}
Dorin Comaniciu and Peter Meer.
\newblock Mean shift: A robust approach toward feature space analysis.
\newblock \emph{PAMI}, 2002.

\bibitem[Curless and Levoy(1996)]{curless1996volumetric}
Brian Curless and Marc Levoy.
\newblock A volumetric method for building complex models from range images.
\newblock In \emph{Conference on Computer graphics and interactive techniques},
  1996.

\bibitem[Dai et~al.(2017)Dai, Chang, Savva, Halber, Funkhouser, and
  Nie{\ss}ner]{dai2017scannet}
Angela Dai, Angel~X Chang, Manolis Savva, Maciej Halber, Thomas Funkhouser, and
  Matthias Nie{\ss}ner.
\newblock {ScanNet}: Richly-annotated {3D} reconstructions of indoor scenes.
\newblock In \emph{CVPR}, 2017.

\bibitem[Dai et~al.(2018)Dai, Ritchie, Bokeloh, Reed, Sturm, and
  Nie{\ss}ner]{dai2018scancomplete}
Angela Dai, Daniel Ritchie, Martin Bokeloh, Scott Reed, J{\"u}rgen Sturm, and
  Matthias Nie{\ss}ner.
\newblock Scancomplete: Large-scale scene completion and semantic segmentation
  for {3D} scans.
\newblock In \emph{CVPR}, 2018.

\bibitem[Dosovitskiy et~al.(2021)Dosovitskiy, Beyer, Kolesnikov, Weissenborn,
  Zhai, Unterthiner, Dehghani, Minderer, Heigold, Gelly,
  et~al.]{dosovitskiy2020image}
Alexey Dosovitskiy, Lucas Beyer, Alexander Kolesnikov, Dirk Weissenborn,
  Xiaohua Zhai, Thomas Unterthiner, Mostafa Dehghani, Matthias Minderer, Georg
  Heigold, Sylvain Gelly, et~al.
\newblock An image is worth 16x16 words: Transformers for image recognition at
  scale.
\newblock In \emph{ICLR}, 2021.

\bibitem[Duzceker et~al.(2021)Duzceker, Galliani, Vogel, Speciale, Dusmanu, and
  Pollefeys]{duzceker2021deepvideomvs}
Arda Duzceker, Silvano Galliani, Christoph Vogel, Pablo Speciale, Mihai
  Dusmanu, and Marc Pollefeys.
\newblock Deepvideomvs: Multi-view stereo on video with recurrent
  spatio-temporal fusion.
\newblock In \emph{CVPR}, 2021.

\bibitem[Fishler(1981)]{fishler1981ransac}
Martin~A Fishler.
\newblock Random sample consensus: A paradigm for model fitting with
  applications to image analysis and automated cartography.
\newblock \emph{Communications of the ACM}, 1981.

\bibitem[Gallup et~al.(2007)Gallup, Frahm, Mordohai, Yang, and
  Pollefeys]{gallup2007real}
David Gallup, Jan-Michael Frahm, Philippos Mordohai, Qingxiong Yang, and Marc
  Pollefeys.
\newblock Real-time plane-sweeping stereo with multiple sweeping directions.
\newblock In \emph{CVPR}, 2007.

\bibitem[Gallup et~al.(2010)Gallup, Frahm, and Pollefeys]{gallup2010piecewise}
David Gallup, Jan-Michael Frahm, and Marc Pollefeys.
\newblock Piecewise planar and non-planar stereo for urban scene
  reconstruction.
\newblock In \emph{CVPR}, 2010.

\bibitem[Geneva et~al.(2018)Geneva, Eckenhoff, Yang, and Huang]{geneva2018lips}
Patrick Geneva, Kevin Eckenhoff, Yulin Yang, and Guoquan Huang.
\newblock Lips: Lidar-inertial {3D} plane {SLAM}.
\newblock In \emph{IROS}, 2018.

\bibitem[Google(2023)]{arcore}
Google.
\newblock {ARCore}, 2023.
\newblock {Accessed: 5 October 2023}.

\bibitem[Hough(1962)]{hough1962method}
Paul~VC Hough.
\newblock Method and means for recognizing complex patterns, 1962.
\newblock US Patent 3,069,654.

\bibitem[Hu et~al.(2022)Hu, Duan, Zhang, Sun, and Huang]{hu2022mvlayoutnet}
Zhihua Hu, Bo Duan, Yanfeng Zhang, Mingwei Sun, and Jingwei Huang.
\newblock {MVLayoutNet: 3D} layout reconstruction with multi-view panoramas.
\newblock In \emph{International Conference on Multimedia}, 2022.

\bibitem[Huang et~al.(2018)Huang, Matzen, Kopf, Ahuja, and
  Huang]{huang2018deepmvs}
Po-Han Huang, Kevin Matzen, Johannes Kopf, Narendra Ahuja, and Jia-Bin Huang.
\newblock Deepmvs: Learning multi-view stereopsis.
\newblock In \emph{CVPR}, 2018.

\bibitem[Hummel et~al.(2006)Hummel, Kammel, Dang, Duchow, and
  Stiller]{hummel2006vision}
Britta Hummel, Soeren Kammel, Thao Dang, Christian Duchow, and Christoph
  Stiller.
\newblock Vision-based path-planning in unstructured environments.
\newblock In \emph{Intelligent Vehicles Symposium}, 2006.

\bibitem[Isack and Boykov(2012)]{isack2012energy}
Hossam Isack and Yuri Boykov.
\newblock Energy-based geometric multi-model fitting.
\newblock \emph{IJCV}, 2012.

\bibitem[Jin et~al.(2021)Jin, Qian, Owens, and Fouhey]{jin2021planar}
Linyi Jin, Shengyi Qian, Andrew Owens, and David~F Fouhey.
\newblock Planar surface reconstruction from sparse views.
\newblock In \emph{ICCV}, 2021.

\bibitem[Kerr et~al.(2023)Kerr, Kim, Goldberg, Kanazawa, and
  Tancik]{kerr2023lerf}
Justin Kerr, Chung~Min Kim, Ken Goldberg, Angjoo Kanazawa, and Matthew Tancik.
\newblock Lerf: Language embedded radiance fields.
\newblock In \emph{ICCV}, 2023.

\bibitem[Kluger et~al.(2021)Kluger, Ackermann, Brachmann, Yang, and
  Rosenhahn]{kluger2021cuboids}
Florian Kluger, Hanno Ackermann, Eric Brachmann, Michael~Ying Yang, and Bodo
  Rosenhahn.
\newblock Cuboids revisited: Learning robust {3D} shape fitting to single {RGB}
  images.
\newblock In \emph{CVPR}, 2021.

\bibitem[Kobayashi et~al.(2022)Kobayashi, Matsumoto, and
  Sitzmann]{kobayashi2022decomposing}
Sosuke Kobayashi, Eiichi Matsumoto, and Vincent Sitzmann.
\newblock Decomposing nerf for editing via feature field distillation.
\newblock In \emph{NeurIPS}, 2022.

\bibitem[Kuhn(1955)]{kuhn1955hungarian}
Harold~W Kuhn.
\newblock The {Hungarian} method for the assignment problem.
\newblock \emph{Naval research logistics quarterly}, 1955.

\bibitem[Kundu et~al.(2022)Kundu, Genova, Yin, Fathi, Pantofaru, Guibas,
  Tagliasacchi, Dellaert, and Funkhouser]{kunduCVPR2022PNF}
Abhijit Kundu, Kyle Genova, Xiaoqi Yin, Alireza Fathi, Caroline Pantofaru,
  Leonidas Guibas, Andrea Tagliasacchi, Frank Dellaert, and Thomas Funkhouser.
\newblock {Panoptic Neural Fields: A Semantic Object-Aware Neural Scene
  Representation}.
\newblock In \emph{CVPR}, 2022.

\bibitem[Lafarge and Alliez(2013)]{lafarge2013surface}
Florent Lafarge and Pierre Alliez.
\newblock Surface reconstruction through point set structuring.
\newblock In \emph{Computer Graphics Forum}, 2013.

\bibitem[Lafarge et~al.(2012)Lafarge, Keriven, Br{\'e}dif, and
  Vu]{lafarge2012hybrid}
Florent Lafarge, Renaud Keriven, Mathieu Br{\'e}dif, and Hoang-Hiep Vu.
\newblock A hybrid multiview stereo algorithm for modeling urban scenes.
\newblock \emph{PAMI}, 2012.

\bibitem[Lee et~al.(2019)Lee, Han, Ko, and Suh]{lee2019big}
Jin~Han Lee, Myung-Kyu Han, Dong~Wook Ko, and Il~Hong Suh.
\newblock From big to small: Multi-scale local planar guidance for monocular
  depth estimation.
\newblock \emph{arXiv:1907.10326}, 2019.

\bibitem[Li et~al.(2023)Li, M{\"u}ller, Evans, Taylor, Unberath, Liu, and
  Lin]{li2023neuralangelo}
Zhaoshuo Li, Thomas M{\"u}ller, Alex Evans, Russell~H Taylor, Mathias Unberath,
  Ming-Yu Liu, and Chen-Hsuan Lin.
\newblock Neuralangelo: High-fidelity neural surface reconstruction.
\newblock In \emph{CVPR}, 2023.

\bibitem[Liu et~al.(2018)Liu, Yang, Ceylan, Yumer, and
  Furukawa]{liu2018planenet}
Chen Liu, Jimei Yang, Duygu Ceylan, Ersin Yumer, and Yasutaka Furukawa.
\newblock Planenet: Piece-wise planar reconstruction from a single {RGB} image.
\newblock In \emph{CVPR}, 2018.

\bibitem[Liu et~al.(2019{\natexlab{a}})Liu, Kim, Gu, Furukawa, and
  Kautz]{liu2019planercnn}
Chen Liu, Kihwan Kim, Jinwei Gu, Yasutaka Furukawa, and Jan Kautz.
\newblock Plane{RCNN}: {3D} plane detection and reconstruction from a single
  image.
\newblock In \emph{CVPR}, 2019{\natexlab{a}}.

\bibitem[Liu et~al.(2022)Liu, Ji, Bansal, Cai, Yan, Huang, and
  Xu]{liu2022planemvs}
Jiachen Liu, Pan Ji, Nitin Bansal, Changjiang Cai, Qingan Yan, Xiaolei Huang,
  and Yi Xu.
\newblock {PlaneMVS: 3D} plane reconstruction from multi-view stereo.
\newblock In \emph{CVPR}, 2022.

\bibitem[Liu et~al.(2019{\natexlab{b}})Liu, Saito, Chen, and
  Li]{liu2019learning}
Shichen Liu, Shunsuke Saito, Weikai Chen, and Hao Li.
\newblock Learning to infer implicit surfaces without {3D} supervision.
\newblock In \emph{NeurIPS}, 2019{\natexlab{b}}.

\bibitem[Mescheder et~al.(2019)Mescheder, Oechsle, Niemeyer, Nowozin, and
  Geiger]{mescheder2019occupancy}
Lars Mescheder, Michael Oechsle, Michael Niemeyer, Sebastian Nowozin, and
  Andreas Geiger.
\newblock Occupancy networks: Learning {3D} reconstruction in function space.
\newblock In \emph{CVPR}, 2019.

\bibitem[Mildenhall et~al.(2020)Mildenhall, Srinivasan, Tancik, Barron,
  Ramamoorthi, and Ng]{mildenhall2020nerf}
Ben Mildenhall, Pratul~P. Srinivasan, Matthew Tancik, Jonathan~T. Barron, Ravi
  Ramamoorthi, and Ren Ng.
\newblock {NeRF}: Representing scenes as neural radiance fields for view
  synthesis.
\newblock In \emph{ECCV}, 2020.

\bibitem[Mitra et~al.(2013)Mitra, Pauly, Wand, and Ceylan]{mitra2013symmetry}
Niloy~J Mitra, Mark Pauly, Michael Wand, and Duygu Ceylan.
\newblock Symmetry in {3D} geometry: Extraction and applications.
\newblock In \emph{Computer Graphics Forum}, 2013.

\bibitem[Monszpart et~al.(2015)Monszpart, Mellado, Brostow, and
  Mitra]{monszpart2015rapter}
Aron Monszpart, Nicolas Mellado, Gabriel~J Brostow, and Niloy~J Mitra.
\newblock {RAPter}: rebuilding man-made scenes with regular arrangements of
  planes.
\newblock \emph{ACM Transactions on Graphics}, 2015.

\bibitem[Murez et~al.(2020)Murez, van As, Bartolozzi, Sinha, Badrinarayanan,
  and Rabinovich]{murez2020atlas}
Zak Murez, Tarrence van As, James Bartolozzi, Ayan Sinha, Vijay Badrinarayanan,
  and Andrew Rabinovich.
\newblock Atlas: End-to-end {3D} scene reconstruction from posed images.
\newblock In \emph{ECCV}, 2020.

\bibitem[Nan and Wonka(2017)]{nan2017polyfit}
Liangliang Nan and Peter Wonka.
\newblock {PolyFit}: Polygonal surface reconstruction from point clouds.
\newblock In \emph{ICCV}, 2017.

\bibitem[Narita et~al.(2019)Narita, Seno, Ishikawa, and
  Kaji]{narita2019panopticfusion}
Gaku Narita, Takashi Seno, Tomoya Ishikawa, and Yohsuke Kaji.
\newblock Panopticfusion: Online volumetric semantic mapping at the level of
  stuff and things.
\newblock In \emph{IROS}, 2019.

\bibitem[Newcombe et~al.(2011)Newcombe, Izadi, and
  Hilliges]{newcombe2011kinectfusion}
Richard~A Newcombe, Shahram Izadi, and Otmar Hilliges.
\newblock {KinectFusion}: Real-time dense surface mapping and tracking.
\newblock In \emph{UIST}, 2011.

\bibitem[Oechsle et~al.(2021)Oechsle, Peng, and Geiger]{oechsle2021unisurf}
Michael Oechsle, Songyou Peng, and Andreas Geiger.
\newblock {UNISURF}: Unifying neural implicit surfaces and radiance fields for
  multi-view reconstruction.
\newblock In \emph{ICCV}, 2021.

\bibitem[Park et~al.(2019)Park, Florence, Straub, Newcombe, and
  Lovegrove]{park2019deepsdf}
Jeong~Joon Park, Peter Florence, Julian Straub, Richard Newcombe, and Steven
  Lovegrove.
\newblock {DeepSDF}: Learning continuous signed distance functions for shape
  representation.
\newblock In \emph{CVPR}, 2019.

\bibitem[Pathak et~al.(2010)Pathak, Birk, Vaskevicius, Pfingsthorn,
  Schwertfeger, and Poppinga]{pathak2010online}
Kaustubh Pathak, Andreas Birk, Narunas Vaskevicius, Max Pfingsthorn, S{\"o}ren
  Schwertfeger, and Jann Poppinga.
\newblock Online three-dimensional {SLAM} by registration of large planar
  surface segments and closed-form pose-graph relaxation.
\newblock \emph{Journal of Field Robotics}, 2010.

\bibitem[Peng et~al.(2023)Peng, Genova, Jiang, Tagliasacchi, Pollefeys,
  Funkhouser, et~al.]{peng2023openscene}
Songyou Peng, Kyle Genova, Chiyu Jiang, Andrea Tagliasacchi, Marc Pollefeys,
  Thomas Funkhouser, et~al.
\newblock {OpenScene}: {3D} scene understanding with open vocabularies.
\newblock In \emph{CVPR}, 2023.

\bibitem[Radford et~al.(2021)Radford, Kim, Hallacy, Ramesh, Goh, Agarwal,
  Sastry, Askell, Mishkin, Clark, et~al.]{radford2021learning}
Alec Radford, Jong~Wook Kim, Chris Hallacy, Aditya Ramesh, Gabriel Goh,
  Sandhini Agarwal, Girish Sastry, Amanda Askell, Pamela Mishkin, Jack Clark,
  et~al.
\newblock Learning transferable visual models from natural language
  supervision.
\newblock In \emph{ICML}, 2021.

\bibitem[Ramamonjisoa et~al.(2022)Ramamonjisoa, Stekovic, and
  Lepetit]{ramamonjisoa2022monteboxfinder}
Micha{\"e}l Ramamonjisoa, Sinisa Stekovic, and Vincent Lepetit.
\newblock Monteboxfinder: Detecting and filtering primitives to fit a noisy
  point cloud.
\newblock In \emph{ECCV}, 2022.

\bibitem[Rethage et~al.(2018)Rethage, Wald, Sturm, Navab, and
  Tombari]{rethage2018fully}
Dario Rethage, Johanna Wald, Jurgen Sturm, Nassir Navab, and Federico Tombari.
\newblock Fully-convolutional point networks for large-scale point clouds.
\newblock In \emph{ECCV}, 2018.

\bibitem[Rozumnyi et~al.(2023)Rozumnyi, Popov, Maninis, Nie{\ss}ner, and
  Ferrari]{rozumnyi2023estimating}
Denys Rozumnyi, Stefan Popov, Kevis-Kokitsi Maninis, Matthias Nie{\ss}ner, and
  Vittorio Ferrari.
\newblock Estimating generic {3D} room structures from {2D} annotations.
\newblock \emph{NeurIPS}, 2023.

\bibitem[Runz et~al.(2020)Runz, Li, Tang, Ma, Kong, Schmidt, Reid, Agapito,
  Straub, Lovegrove, et~al.]{runz2020frodo}
Martin Runz, Kejie Li, Meng Tang, Lingni Ma, Chen Kong, Tanner Schmidt, Ian
  Reid, Lourdes Agapito, Julian Straub, Steven Lovegrove, et~al.
\newblock {FroDO}: From detections to {3D} objects.
\newblock In \emph{CVPR}, 2020.

\bibitem[Sayed et~al.(2022)Sayed, Gibson, Watson, Prisacariu, Firman, and
  Godard]{sayed2022simplerecon}
Mohamed Sayed, John Gibson, Jamie Watson, Victor Prisacariu, Michael Firman,
  and Cl{\'e}ment Godard.
\newblock Simplerecon: {3D} reconstruction without {3D} convolutions.
\newblock In \emph{ECCV}, 2022.

\bibitem[Schmid et~al.(2022)Schmid, Delmerico, Sch{\"o}nberger, Nieto,
  Pollefeys, Siegwart, and Cadena]{schmid2022panoptic}
Lukas Schmid, Jeffrey Delmerico, Johannes Sch{\"o}nberger, Juan Nieto, Marc
  Pollefeys, Roland Siegwart, and Cesar Cadena.
\newblock Panoptic multi-{TSDFs}: a flexible representation for online
  multi-resolution volumetric mapping and long-term dynamic scene consistency.
\newblock In \emph{ICRA}, 2022.

\bibitem[Schnabel et~al.(2007)Schnabel, Wahl, and Klein]{schnabel2007efficient}
Ruwen Schnabel, Roland Wahl, and Reinhard Klein.
\newblock Efficient ransac for point-cloud shape detection.
\newblock In \emph{Computer Graphics Forum}, 2007.

\bibitem[Seichter et~al.(2023)Seichter, Stephan, Fischedick, Mueller, Rabes,
  and Gross]{panopticndt2023iros}
Daniel Seichter, Benedict Stephan, S{\"o}hnke~Benedikt Fischedick, Steffen
  Mueller, Leonard Rabes, and Horst-Michael Gross.
\newblock {PanopticNDT: Efficient and Robust Panoptic Mapping}.
\newblock In \emph{IROS}, 2023.

\bibitem[Shao et~al.(2023)Shao, Pei, Chen, Wu, and Li]{shao2023nddepth}
Shuwei Shao, Zhongcai Pei, Weihai Chen, Xingming Wu, and Zhengguo Li.
\newblock {NDDepth}: Normal-distance assisted monocular depth estimation.
\newblock In \emph{ICCV}, 2023.

\bibitem[Shi et~al.(2023)Shi, Zhi, and Xu]{shi2023planerectr}
Jingjia Shi, Shuaifeng Zhi, and Kai Xu.
\newblock {PlaneRecTR}: Unified query learning for {3D} plane recovery from a
  single view.
\newblock In \emph{ICCV}, 2023.

\bibitem[Siddiqui et~al.(2023)Siddiqui, Porzi, Bul\`o, M\"uller, Nie{\ss}ner,
  Dai, and Kontschieder]{siddiqui2023panoptic_lift}
Yawar Siddiqui, Lorenzo Porzi, Samuel~Rota Bul\`o, Norman M\"uller, Matthias
  Nie{\ss}ner, Angela Dai, and Peter Kontschieder.
\newblock Panoptic lifting for {3D} scene understanding with neural fields.
\newblock In \emph{CVPR}, 2023.

\bibitem[Sinha et~al.(2008)Sinha, Steedly, Szeliski, Agrawala, and
  Pollefeys]{sinha2008interactive}
Sudipta~N Sinha, Drew Steedly, Richard Szeliski, Maneesh Agrawala, and Marc
  Pollefeys.
\newblock Interactive {3D} architectural modeling from unordered photo
  collections.
\newblock \emph{Transactions on Graphics}, 2008.

\bibitem[Sitzmann et~al.(2020)Sitzmann, Martel, Bergman, Lindell, and
  Wetzstein]{sitzmann2020implicit}
Vincent Sitzmann, Julien Martel, Alexander Bergman, David Lindell, and Gordon
  Wetzstein.
\newblock Implicit neural representations with periodic activation functions.
\newblock \emph{NeurIPS}, 2020.

\bibitem[Stier et~al.(2023)Stier, Ranjan, Colburn, Yan, Yang, Ma, and
  Angles]{stier2023finerecon}
Noah Stier, Anurag Ranjan, Alex Colburn, Yajie Yan, Liang Yang, Fangchang Ma,
  and Baptiste Angles.
\newblock {FineRecon}: Depth-aware feed-forward network for detailed {3D}
  reconstruction.
\newblock \emph{arXiv:2304.01480}, 2023.

\bibitem[Su et~al.(2023)Su, Peng, Wonka, and Chu]{su2023gpr}
Jheng-Wei Su, Chi-Han Peng, Peter Wonka, and Hung-Kuo Chu.
\newblock {GPr-Net}: Multi-view layout estimation via a geometry-aware panorama
  registration network.
\newblock In \emph{CVPR}, 2023.

\bibitem[Sucar et~al.(2021)Sucar, Liu, Ortiz, and Davison]{sucar2021imap}
Edgar Sucar, Shikun Liu, Joseph Ortiz, and Andrew~J Davison.
\newblock {iMAP}: Implicit mapping and positioning in real-time.
\newblock In \emph{ICCV}, 2021.

\bibitem[Sun et~al.(2021)Sun, Xie, Chen, Zhou, and Bao]{sun2021neuralrecon}
Jiaming Sun, Yiming Xie, Linghao Chen, Xiaowei Zhou, and Hujun Bao.
\newblock {NeuralRecon}: Real-time coherent {3D} reconstruction from monocular
  video.
\newblock In \emph{CVPR}, 2021.

\bibitem[Tan et~al.(2021)Tan, Xue, Bai, Wu, and Xia]{tan2021planeTR}
Bin Tan, Nan Xue, Song Bai, Tianfu Wu, and Gui-Song Xia.
\newblock {PlaneTR}: Structure-guided transformers for {3D} plane recovery.
\newblock In \emph{ICCV}, 2021.

\bibitem[Tan et~al.(2023)Tan, Xue, Wu, and Xia]{tan2023nope}
Bin Tan, Nan Xue, Tianfu Wu, and Gui-Song Xia.
\newblock {NOPE-SAC}: Neural one-plane {RANSAC} for sparse-view planar {3D}
  reconstruction.
\newblock \emph{PAMI}, 2023.

\bibitem[Tsagkas et~al.(2023)Tsagkas, Mac~Aodha, and Lu]{tsagkas2023vl}
Nikolaos Tsagkas, Oisin Mac~Aodha, and Chris~Xiaoxuan Lu.
\newblock Vl-fields: Towards language-grounded neural implicit spatial
  representations.
\newblock In \emph{ICRA Workshops}, 2023.

\bibitem[Turner and Zakhor(2013)]{turner2013watertight}
Eric Turner and Avideh Zakhor.
\newblock Watertight planar surface meshing of indoor point-clouds with voxel
  carving.
\newblock In \emph{3DV}, 2013.

\bibitem[Uematsu and Saito(2009)]{uematsu2009multiple}
Yuko Uematsu and Hideo Saito.
\newblock Multiple planes based registration using {3D} projective space for
  augmented reality.
\newblock \emph{Image and Vision Computing}, 2009.

\bibitem[Vanegas et~al.(2012)Vanegas, Aliaga, and Benes]{vanegas2012automatic}
Carlos~A Vanegas, Daniel~G Aliaga, and Bedrich Benes.
\newblock Automatic extraction of {Manhattan}-world building masses from {3D}
  laser range scans.
\newblock \emph{Transactions on Visualization and Computer Graphics}, 2012.

\bibitem[Wang et~al.(2021)Wang, Liu, Liu, Theobalt, Komura, and
  Wang]{wang2021neus}
Peng Wang, Lingjie Liu, Yuan Liu, Christian Theobalt, Taku Komura, and Wenping
  Wang.
\newblock {NeuS}: Learning neural implicit surfaces by volume rendering for
  multi-view reconstruction.
\newblock In \emph{NeurIPS}, 2021.

\bibitem[Xia et~al.(2020)Xia, Chen, Wang, Li, and Zhang]{xia2020geometric}
Shaobo Xia, Dong Chen, Ruisheng Wang, Jonathan Li, and Xinchang Zhang.
\newblock Geometric primitives in lidar point clouds: A review.
\newblock \emph{Journal of Selected Topics in Applied Earth Observations and
  Remote Sensing}, 2020.

\bibitem[Xie et~al.(2022)Xie, Gadelha, Yang, Zhou, and
  Jiang]{xie2022planarrecon}
Yiming Xie, Matheus Gadelha, Fengting Yang, Xiaowei Zhou, and Huaizu Jiang.
\newblock {PlanarRecon}: Real-time {3D} plane detection and reconstruction from
  posed monocular videos.
\newblock In \emph{CVPR}, 2022.

\bibitem[Yang and Zhou(2018)]{yang2018recovering}
Fengting Yang and Zihan Zhou.
\newblock Recovering {3D} planes from a single image via convolutional neural
  networks.
\newblock In \emph{ECCV}, 2018.

\bibitem[Ye et~al.(2023)Ye, Liu, Li, and Yang]{ye2023self}
Botao Ye, Sifei Liu, Xueting Li, and Ming-Hsuan Yang.
\newblock Self-supervised super-plane for neural {3D} reconstruction.
\newblock In \emph{CVPR}, 2023.

\bibitem[Yu et~al.(2019)Yu, Zheng, Lian, Zhou, and Gao]{yu2019single}
Zehao Yu, Jia Zheng, Dongze Lian, Zihan Zhou, and Shenghua Gao.
\newblock Single-image piece-wise planar {3D} reconstruction via associative
  embedding.
\newblock In \emph{CVPR}, 2019.

\bibitem[Yu et~al.(2020)Yu, Jin, and Gao]{yu2020p}
Zehao Yu, Lei Jin, and Shenghua Gao.
\newblock P$^2$net: Patch-match and plane-regularization for unsupervised
  indoor depth estimation.
\newblock In \emph{ECCV}, 2020.

\bibitem[Yu et~al.(2022)Yu, Peng, Niemeyer, Sattler, and Geiger]{yu2022monosdf}
Zehao Yu, Songyou Peng, Michael Niemeyer, Torsten Sattler, and Andreas Geiger.
\newblock {MonoSDF}: Exploring monocular geometric cues for neural implicit
  surface reconstruction.
\newblock \emph{NeurIPS}, 2022.

\bibitem[Zhang et~al.(2023)Zhang, Tosi, Mattoccia, and Poggi]{zhang2023goslam}
Youmin Zhang, Fabio Tosi, Stefano Mattoccia, and Matteo Poggi.
\newblock {GO-SLAM}: Global optimization for consistent {3D} instant
  reconstruction.
\newblock In \emph{ICCV}, 2023.

\bibitem[Zhi et~al.(2021{\natexlab{a}})Zhi, Laidlow, Leutenegger, and
  Davison]{zhi2021inplace}
Shuaifeng Zhi, Tristan Laidlow, Stefan Leutenegger, and Andrew Davison.
\newblock In-place scene labelling and understanding with implicit scene
  representation.
\newblock In \emph{ICCV}, 2021{\natexlab{a}}.

\bibitem[Zhi et~al.(2021{\natexlab{b}})Zhi, Sucar, Mouton, Haughton, Laidlow,
  and Davison]{zhi2021ilabel}
Shuaifeng Zhi, Edgar Sucar, Andr{\'{e}} Mouton, Iain Haughton, Tristan Laidlow,
  and Andrew~J. Davison.
\newblock {iLabel}: Revealing objects in neural fields.
\newblock \emph{Robotics and Automation Letters}, 2021{\natexlab{b}}.

\bibitem[Zhu et~al.(2022)Zhu, Peng, Larsson, Xu, Bao, Cui, Oswald, and
  Pollefeys]{zhu2022niceslam}
Zihan Zhu, Songyou Peng, Viktor Larsson, Weiwei Xu, Hujun Bao, Zhaopeng Cui,
  Martin~R. Oswald, and Marc Pollefeys.
\newblock {NICE-SLAM}: Neural implicit scalable encoding for {SLAM}.
\newblock In \emph{CVPR}, 2022.

\end{thebibliography}
}

\end{document}